%% file: main.tex
\documentclass[10pt,twocolumn,letterpaper]{article}

\usepackage{cvpr}
\usepackage{times}
\usepackage{tabu}
\usepackage{epsfig}
\usepackage{graphicx}
\usepackage{subcaption}
\usepackage[export]{adjustbox}
\usepackage{amsmath}
\usepackage{amssymb}
\usepackage{algorithmic}
\usepackage{xcolor}
\usepackage[percent]{overpic}
\usepackage{array}

\usepackage[pagebackref=true,breaklinks=true,letterpaper=true,colorlinks,bookmarks=false]{hyperref}

\input{defs}

\cvprfinalcopy 

\def\cvprPaperID{542} 

\ifcvprfinal\fi
\begin{document}

\title{AutoScaler: Scale-Attention Networks for Visual Correspondence}

\author{Shenlong Wang \\
{\small University of Toronto} \\
{\tt\footnotesize slwang@cs.toronto.edu}
\and Linjie Luo \\
{\small Snap Inc.} \\
{\tt\footnotesize linjie.luo@snap.com}
\and Ning Zhang \\
{\small Snap Inc.} \\
{\tt\footnotesize ning.zhang@snap.com}
\and Li-Jia Li \\
{\small Snap Inc.} \\
{\tt\footnotesize lijiali@cs.stanford.edu}
}

\maketitle

\input{abstract}
\input{intro}
\input{related}

\input{method}
\input{exp}

\input{conclusion}

{\small
\bibliographystyle{ieee}
\bibliography{egbib}
}

\end{document}

%% file: defs.tex
\newcommand{\cN}{\mathcal{N}}

\newcommand{\bw}{\mathbf{w}}
\newcommand{\bp}{\mathbf{p}} 
\newcommand{\bq}{\mathbf{q}}

\newcommand{\bx}{\mathbf{x}} 

%% file: abstract.tex
\begin{abstract}
Finding visual correspondence between local features is key to many computer vision problems. While defining features with larger contextual scales usually implies greater discriminativeness, it could also lead to less spatial accuracy of the features. We propose AutoScaler, a scale-attention network to explicitly optimize this trade-off in visual correspondence tasks. Our network consists of a weight-sharing feature network to compute multi-scale feature maps and an attention network to combine them optimally in the scale space. This allows our network to have adaptive receptive field sizes over different scales of the input. The entire network is trained end-to-end in a siamese framework for visual correspondence tasks. Our method achieves favorable results compared to state-of-the-art methods on challenging optical flow and semantic matching benchmarks, including Sintel, KITTI and CUB-2011. We also show that our method can generalize to improve hand-crafted descriptors (e.g Daisy) on general visual correspondence tasks. Finally, our attention network can generate visually interpretable scale attention maps. 

\end{abstract}

%% file: intro.tex
\section{Introduction}
\begin{figure*}
\begin{center}
\includegraphics[width=0.95\linewidth]{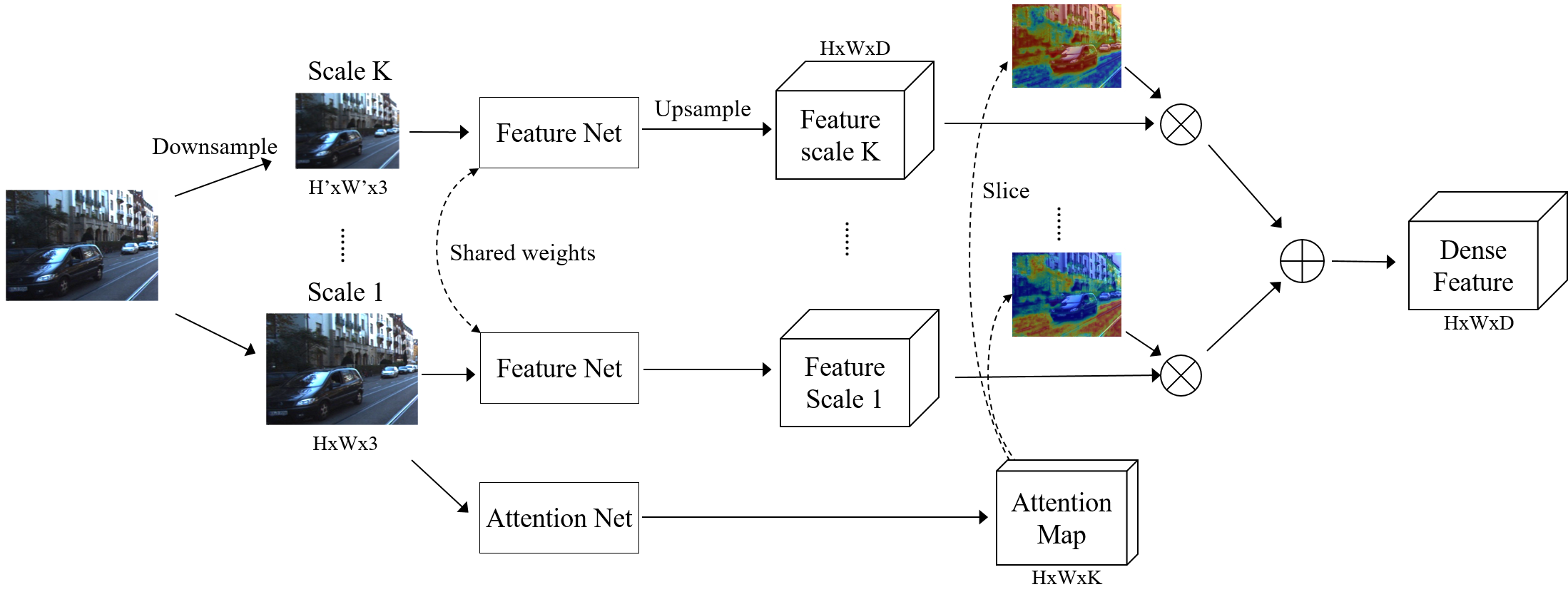}
\end{center}
  \vspace{-5mm}
\caption{The architecture of AutoScaler. AutoScaler consists of the feature network and the attention network. The feature network extracts feature maps from the input image at multiple scales (note that only two are shown for simplicity) independently with shared weights. The separate attention network computes a pixel-wise attention map from the input image and combine the multi-scale feature maps into one.}
  \label{fig:architecture}
  \vspace{-3mm}
\end{figure*}

Finding correspondences between local features in multiple related images is a fundamental problem in computer vision. It is crucial for a plethora of applications, including optical flow \cite{gpc, epicflow, discreteflow}, structure-from-motion \cite{rome}, visual SLAM \cite{dtam, libviso2,orbslam}, stereo matching \cite{mccnn,deepstereo}, non-rigid 3D reconstruction \cite{fusion4d} as well as video segmentation \cite{videoseg}. 

Central to the correspondence problem is the design of feature descriptors that needs to be resilient to lighting change and different object poses and scales. To select the characteristic scales, many hand-crafted descriptors analyze feature saliency in a scale space formed by applying heuristic image processing operators on different scales of the images. The resultant descriptors are extracted from either one \cite{sift,surf} or many \cite{sifting} of these scales. However, due to their heuristic nature, the scale analyses of these hand-crafted descriptors are limited to a sparse set of image locations with special structures, such as blobs, corners and high contrast regions \cite{affine-region}. To compute dense correspondences using these descriptors, one needs to impose smoothness prior to regularize the correspondence map from the sparse matches, which often experiences loss in accuracy \cite{SGM, BroxMalik, PatchMatch}.

Recently, convolutional neural network (CNN) triumphed in a variety of challenging computer vision tasks such as image classification \cite{alexnet,googlenet,resnet} and object detection \cite{rcnn,fasterrcnn}. What makes CNN powerful is its flexible architecture to learn progressively complex visual features from low-level filters to high-level concepts. While higher-level features prove to be discriminative for many applications, they often come with a loss in spatial accuracy in the process of yielding larger receptive fields through successive pooling \cite{alexnet,vgg}, large strides \cite{alexnet}, dilated convolution \cite{dilation} and multi-scale aggregation \cite{googlenet, fcn}. In applications that require spatially accurate correspondences, techniques such as spatial transformer network \cite{ucn} and multi-scale ensemble model \cite{baidu-deep} prove effective to improve discriminativeness while keeping the spatial accuracy in the resultant feature maps. However, further study is needed on how to optimally combine features from different scales based on the analysis in the scale space.

In this paper, we propose the \emph{AutoScaler}, a scale-attention network to optimally combine feature maps from different scales for visual correspondence tasks. Our key insight is that the trade-off between the spatial accuracy and the discriminative contextual scales of local features can be explicitly optimized via a scale-attention network to improve visual correspondence accuracy. Specifically, in texture-rich area, the network will weigh more on the fine-scale features to ensure correspondence accuracy while in area with less texture, the network will seek for the features at larger scales for more discriminative contextual information.

Our AutoScaler network consists of a weight-sharing \emph{feature network} to compute multi-scale feature maps and an \emph{attention network} to combine them optimally in the scale space (Fig. \ref{fig:architecture}). By sharing weights across multi-scale feature network, our system can handle large scale changes. The proposed network can generalize to improve the performance of handcrafted descriptors (e.g. Daisy \cite{daisy}). The full network is trained end-to-end in a siamese framework (Fig. \ref{fig:inference}) without explicit supervision on scale-attention. We demonstrate the effectiveness of the proposed method over optical flow, semantic correspondence tasks and find it compared favorably with the state-of-the-art methods. Moreover, our algorithm is able to generate visually interpretable scale attention maps.

%% file: related.tex
\section{Related Work}

Our work is closely related to learning based approaches for image correspondence. Early methods consider image correspondence problem as a variational inference and focus on learn parameters for MRFs \cite{foeflow}. Later on, the growing availability of synthetic and real world datasets for image correspondence problems makes learning feature representation for similarity matching possible. Representative works include the usage of boosting \cite{boosting}, random forest \cite{gpc}, convex optimization \cite{convex}, \etc. Recently multiple CNNs based approaches are designed to measure similarity between patches across images \cite{matchnet,ning,deepstereo,mccnn, baidu-deep, siamese, warpnet, universalnet}. In particular, our method is related to \cite{deepstereo} in terms of loss functions.

Scale selection has been extensively studied in previous work \cite{sift,mser,scale-selection} to select the most salient scale for matching in the scale space. However, the analyses are limited to a few heuristic rules that apply to a sparse set of key points. 
It is non-trivial to extend the scale selection for dense matching problems such as optical flow.
Instead, many previous works explore to propagate the scale labels from sparse key points to the whole image \cite{sifting,scalesift,sls}. But it imposes strong smoothness prior which leads to degraded matching accuracy.

Many approaches have been proposed to enlarge receptive field size to incorporate more contextual information, such as dilated convolution \cite{dilation}, multi-scale aggregation \cite{googlenet, fcn}, pooling \cite{alexnet} and large strides \cite{alexnet}. Despite greater discriminativeness, we argue that larger receptive field is not always better for correspondence tasks because it often `blurs' the feature map and reduces pixel-level spatial accuracy as supported by previous experiments \cite{mccnn,siamese,deepstereo}. 
Thus, we propose to optimize the trade-off between larger contextual scale and spatial accuracy by using an scale attention scheme.

Attention mechanism in neural network has been studied in \cite{attention_nips14,attention_icml15,draw15, ba-attention-2015} with impressive results for different computer vision tasks.
The proposed scale-attention model is most related to scale-attention based semantic segmentation method \cite{chen} with two main differences. First, our goal is to find the best trade-off between discriminative contextual scale and spatial accuracy, while \cite{chen} 
aims at handling meaningful semantic objects with different sizes. Second, our proposed scale-attention mechanism puts attentions over multi-scale features for matching, whereas \cite{chen} utilizes the attention as late-fusion weights over the predicted output from multiple scales.


%% file: method.tex
\section{Method}

\begin{figure*}
\begin{center}
\includegraphics[width=0.98\linewidth]{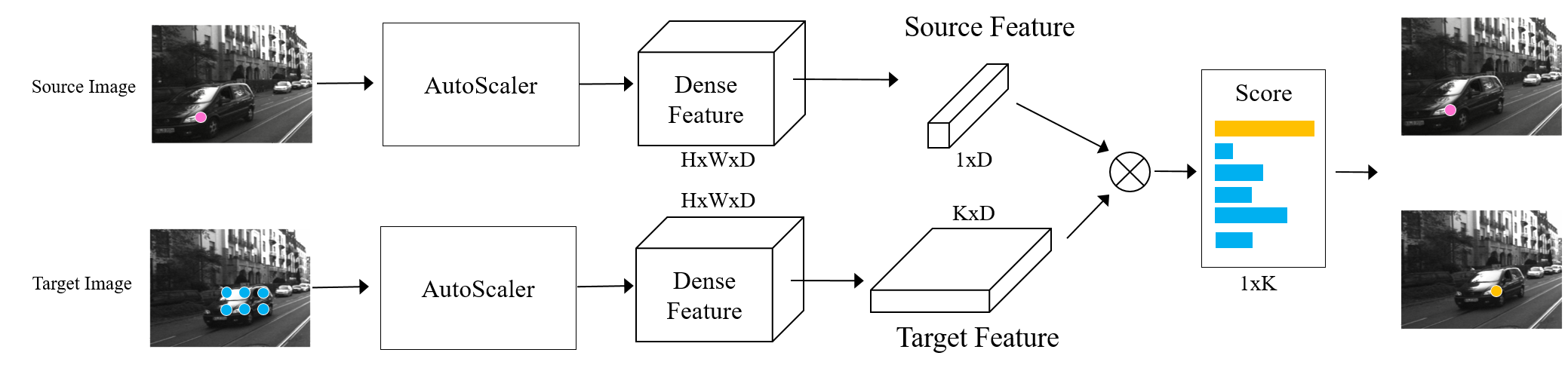}
\end{center}
  \vspace{-3mm}
\caption{The siamese architecture for visual correspondence. Both source image and target image are fed to our AutoScaler network to extract feature maps. One source feature and a number of target features are selected based on the task at hand. Finally, an inner-product layer is employed to find the correspondence with the best score.}
  \label{fig:inference}
\end{figure*}

In this section, we will elaborate our formulation for the visual correspondence tasks of interest as well as implementation details to train the underlying models.

\subsection{Formulation}
We are interested in finding distinctive local correspondence given a pair of related images $I$ and $I^\prime$. A typical correspondence problem tackles the problem by computing a similarity measure $s(\mathbf{p}_i, \mathbf{q}_j)$ between a given position $\mathbf{p}_i$ from the source image and its all possible matching candidates $\cN_\bp = \{\mathbf{q}_{j| j = 1, ..., N}\}$ in the target images; and choose the most similar sample. The candidates set $\cN_\bp$ varies depending on tasks. For instance, we search points along the epipolar line for stereo matching, within a 2D neighborhood for optical flow, and within the whole image for semantic matching. Computation of the similarity measure is typically done by measuring the cost associated with local features located at $\bp$ and $\bq$. 


\begin{figure}[t]
\begin{subfigure}[t]{0.24\linewidth}
 \includegraphics[width=\textwidth]{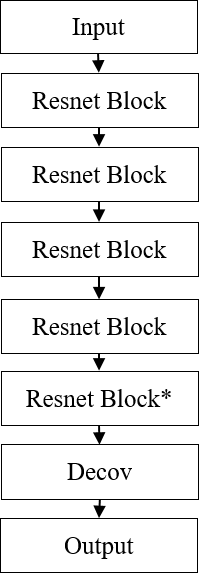}
 \caption{Feature net}
 \end{subfigure}
 ~
 \begin{subfigure}[t]{0.265\linewidth}
 \includegraphics[width=\textwidth]{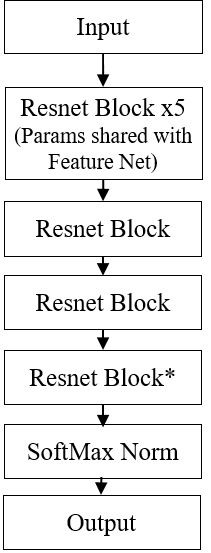}
 \caption{Attention net}
 \end{subfigure}
~
\begin{subfigure}[t]{0.35\linewidth}
 \includegraphics[width=\textwidth]{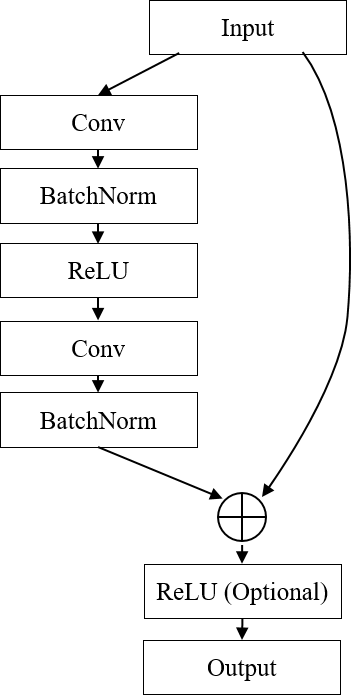}
 \caption{One ResNet block}
 \end{subfigure}
\caption{The detailed architecture of the feature network and the attention network. Both networks are built from basic ResNet blocks. $^\ast$Note that the last ResNet blocks of attention / feature networks do not have ReLU layer.}
  \label{fig:speci_architecture}
\end{figure}

Our general matching scheme is a siamese architecture shown in Fig.~\ref{fig:inference}, where each branch processes the source and target images separately with sharing parameters. In the feature extraction stage, 
each image is passed into a scale-attention network, called AutoScaler. AutoScaler firstly generates a pyramid of input images across different scales as shown in Fig.~\ref{fig:architecture}. Each scale is then passed into a CNN feature net and produces a feature map. The parameters of CNN feature net are shared, which makes same input image 
across multiple scales generate correlated features. Each scale's output is upsampled into the original size of the input image, in order to ensure that the feature maps across scales have the same size. In the meantime, an attention network is introduced to predict a dense weight map for each point across all the scales. The final dense feature is then computed through a weighted sum across all the scales. Fig.~\ref{fig:architecture} depicts the whole process of the dense scale-aware feature computation.

In the matching stage, after we get the dense feature maps, for each point that we are interested in from the source image, we extract its corresponding source feature as well as the features from all the candidate points in the target image. Then an inner-product layer is used to generate the similarity between the source feature and the target features. Point with highest similarity is picked as a corresponding point in the target image. Fig.~\ref{fig:inference} depicts the detailed inference process. 

\begin{figure*}[t]
\definecolor{color-scale1}{RGB}{252,103,105}
\definecolor{color-scale2}{RGB}{254,230,74}
\definecolor{color-scale3}{RGB}{90,253,137}
\definecolor{color-scale4}{RGB}{176,74,251}
\centering
\def\imw{0.245\textwidth}
\setlength{\tabcolsep}{1pt}
\begin{tabular}{cccc}
\includegraphics[width=\imw]{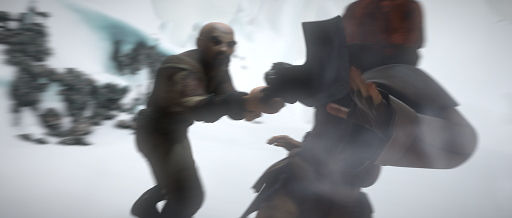} &
\includegraphics[width=\imw]{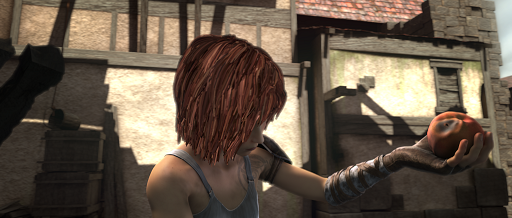} &
\includegraphics[width=\imw]{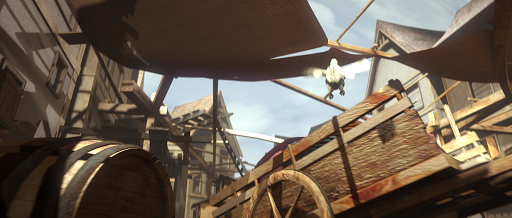} &
\includegraphics[width=\imw]{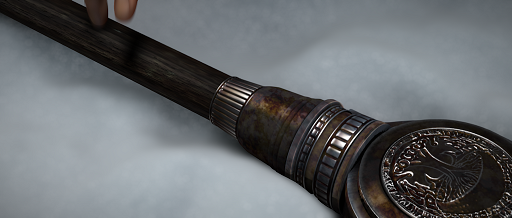} \\
[-2pt]
\includegraphics[width=\imw]{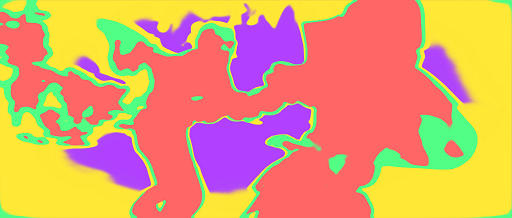} &
\includegraphics[width=\imw]{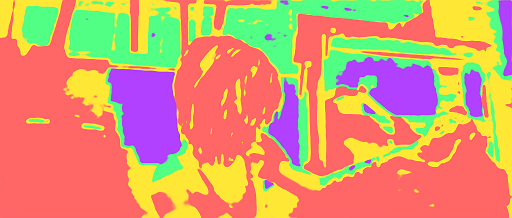} &
\includegraphics[width=\imw]{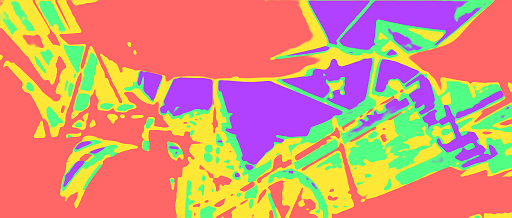} &
\includegraphics[width=\imw]{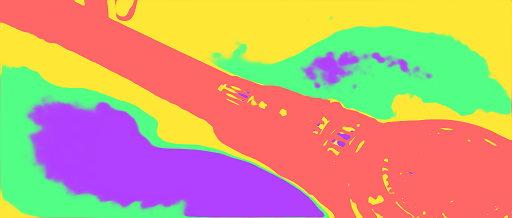} \\ 
[+1pt]
\includegraphics[width=\imw]{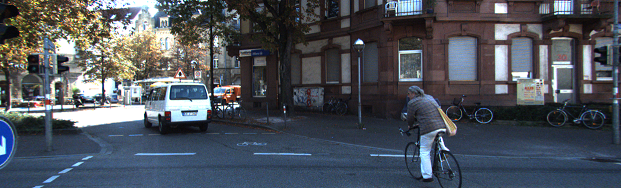} &
\includegraphics[width=\imw]{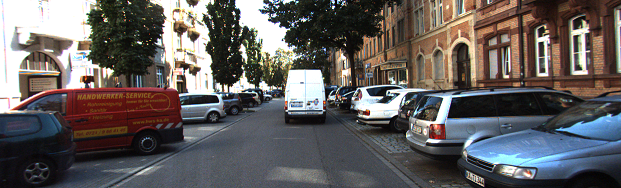} &
\includegraphics[width=\imw]{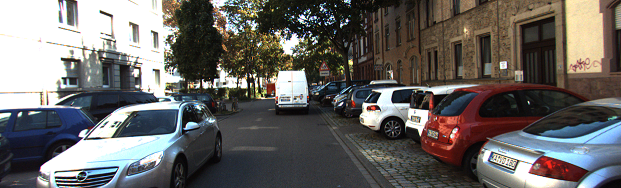} &
\includegraphics[width=\imw]{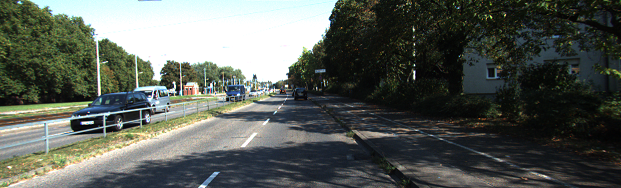} \\
[-2pt]
\includegraphics[width=\imw]{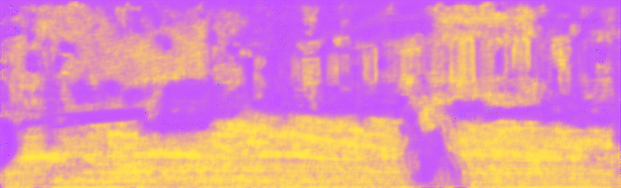} &
\includegraphics[width=\imw]{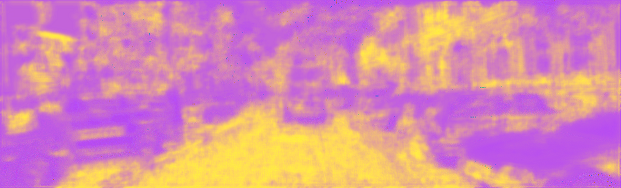} &
\includegraphics[width=\imw]{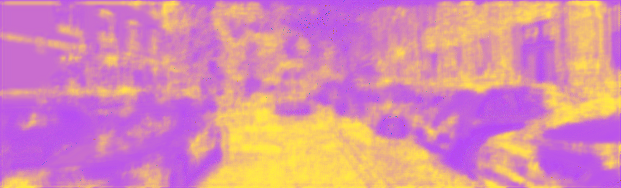} &
\includegraphics[width=\imw]{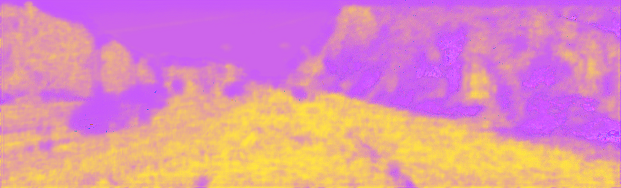} \\
\multicolumn{4}{c}{
{\textcolor{color-scale1}{\rule{1cm}{0.2cm}} scale 0.5$\quad\quad$}
{\textcolor{color-scale2}{\rule{1cm}{0.2cm}} scale 1$\quad\quad$}
{\textcolor{color-scale3}{\rule{1cm}{0.2cm}} scale 1.5$\quad\quad$}
{\textcolor{color-scale4}{\rule{1cm}{0.2cm}} scale 2}
}\\
[-6pt]
\end{tabular}
\label{fig:sintel_attention}
\caption{Visualization of scale-attention maps. Four scales are shown for Sintel dataset (top) and two for KITTI (bottom). Scale values indicate down-sample factors of the input image to the feature network (0.5 being the up-sampled finest scale and 2 the down-sampled coarsest). Note that the attention network weighs more on fine scale for texture-rich regions and gradually moves to larger scales in regions with less texture. }
\end{figure*}
 
\paragraph{Architecture} 
Both the attention network and feature network have a fully convolutional network architecture with shortcut connections to generate pixel-wise feature/score map. 
The CNN feature net contains five ResNet \cite{resnet} blocks, each of which contains a \texttt{conv}-\texttt{batchnorm}-\texttt{relu}-\texttt{conv}-\texttt{batchnorm} structure, followed by a short-cut element-wise sum and a final \texttt{relu} layer. The last \texttt{relu} unit in the feature net is removed for non-sparse feature map. The attention map contains 9 ResNet blocks, with top five sharing parameters with the feature net. The final output is passed through a pixel-wise soft-max layer to ensure the attention weights lies on the range $[0, 1]$ with interpretability. 
Note that we do not use any pooling or strided convolution to ensure that feature/score maps preserve sub-pixel level information. 
Thus the receptive field size is equal to $23 \times 23$ for a single scale feature net. Please refer Fig.~\ref{fig:speci_architecture} for an illustration. In our experiments, the number of filters is 64 (sintel and CUB) or 128 (KITTI). In the following sections, we use 64 filters as example to describe our method.

\begin{table*}[t]
\begin{center}
\begin{tabu}{c|ccccccc} \hline
Dataset & Daisy concat & Daisy attention & CNN-31x31 \cite{deepstereo} & Single & Concat x2 &  AutoScaler x2 & AutoScaler x4 \\ \hline
Sintel & 56.79\%  & 78.30\% & 86.02\% & 86.95\% & 87.65\% & 89.12\% & \textbf{91.84\%} \\ 
KITTI & 73.63\%  & 75.67\% & 90.10\% & 90.07\% & 88.04\% & \textbf{92.06\%} & 91.78\%\\ 
\hline
\end{tabu}
\end{center}
\vspace{-5mm}
\caption{Top-1 accuracy over validation dataset. Our proposed AutoScaler outperforms all the baseline networks. }
\label{tab:top1}
\end{table*}%

\subsection{Training}

\paragraph{Training data} We use the ground-truth pixel-wise correspondence from the dataset to train our neural network. For each pair of images we pick a subset of corresponding pixel pairs. For each pair in the target images, we randomly sample some pixels over all the candidates within the searching range of ground-truth as negative points. This negative sampling is motivated by the fact that points nearby the ground-truth are most likely to be false positive.
In practice we choose 200 negative samples and this results in 201 candidates for each pair with one ground-truth for each source point. We extract features from these points, which results in $64$-dimensional source vector and $64 \times 201$-dimensional target feature. 

\paragraph{Loss} Through computing the inner product between the source feature and all the columns in the target feature, we have a $201$-dimensional score vector describing the confidence of each possible candidates to be a correspondent point. Intuitively, we expect the GT correspondent to have higher score while others have lower score. Thus we minimize cross-entropy loss with respect to the parameters of our neural networks. Let us denote the $i$-th training example a triplet that includes the source feature and all the candidate target features $\bx_i = \{\bp_i, \bq_{j \in \cN_i} \}$, the goal is to
\[
\min_{\bw} \sum_{i, y_i \in \cN_i} p_{\textrm{GT}}(y_i) \log p(y_i; \bx_i, \bw)
\]
where $p(y_i; \bx_i, \bw)$ is the softmax probability $p(y_i; \bx_i, \bw)= \frac{\exp(g_{y_i}(\bx_i, \bw))}{\sum_j \exp(g_{j}(\bx_i, \bw))}$ and the score $g_j(\bx_i, \bw)$ is the inner product between $\bp_i$ and $\bq_j$:
$
g_j(\bx_i, \bw) = \langle \bp_i,  \bq_j \rangle
$; 
The ground-truth probability $p_{\textrm{GT}}(y_i) = 1$ if $y_i$ is GT correspondence and otherwise $p_{\textrm{GT}}(y_i) = 0$. 
where $\bw$ represents all the parameters in the scale-attention network that we want to learn through back-propagation, including both feature net and attention net. 

\paragraph{Optimization} We train our network using stochastic gradient descent with Nesterov momentum. The momentum is set to be 0.9 and the initial learning rate is set to be 0.002. A learning rate policy is set to reduce the learning rate by a factor of 5 for every 50K iterations.

\begin{figure*}
\centering
\def\imw{0.215\textwidth}
\setlength{\tabcolsep}{1pt}
\begin{tabular}{ccccc}
\raisebox{20pt}{Ground-truth} &
\begin{overpic}[width=\imw]{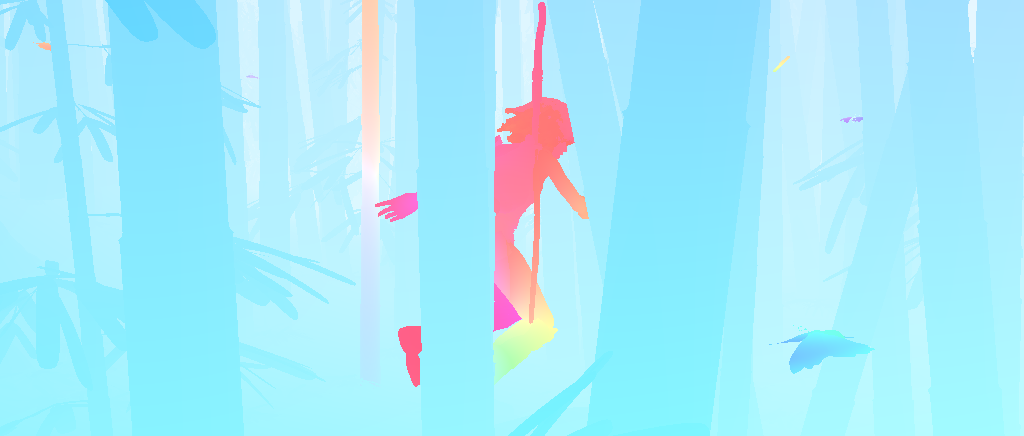}
\put(45,5){\framebox(15,35){}}
\end{overpic} &
\begin{overpic}[width=\imw]{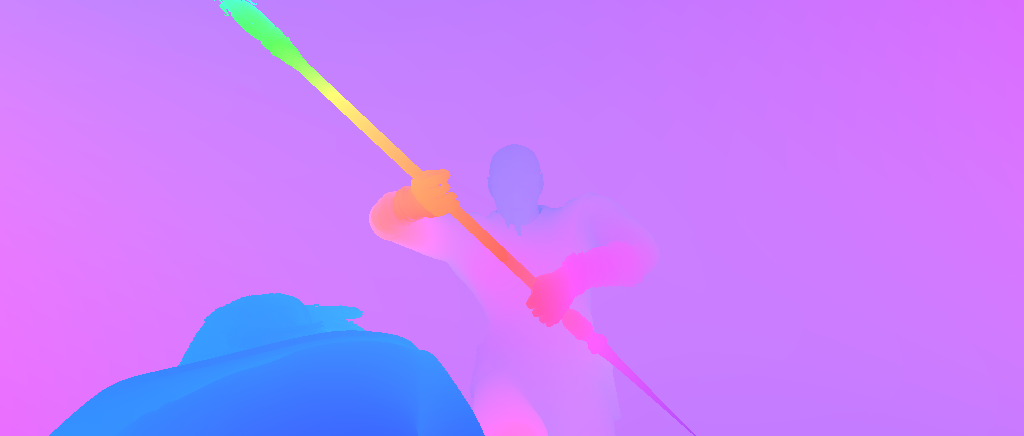}
\put(25,20){\framebox(20,20){}}
\end{overpic} &
\begin{overpic}[width=\imw]{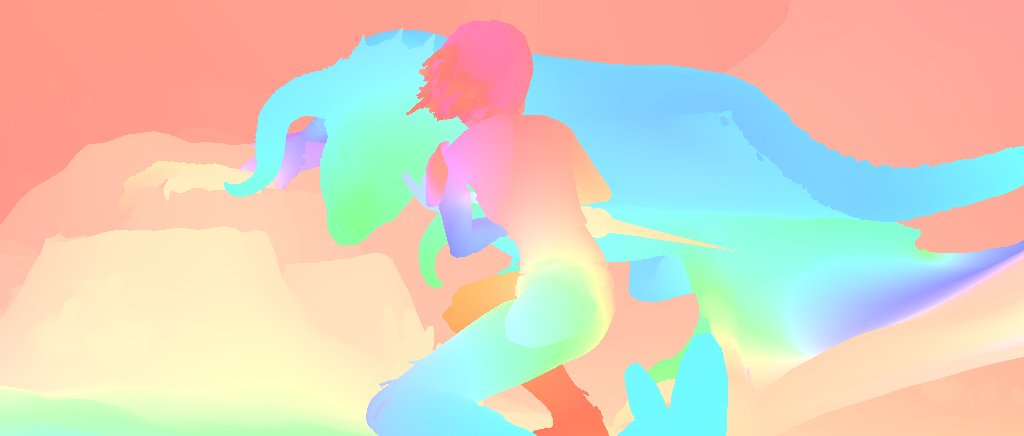}
\put(40,0){\framebox(20,20){}}
\end{overpic} &
\begin{overpic}[width=\imw]{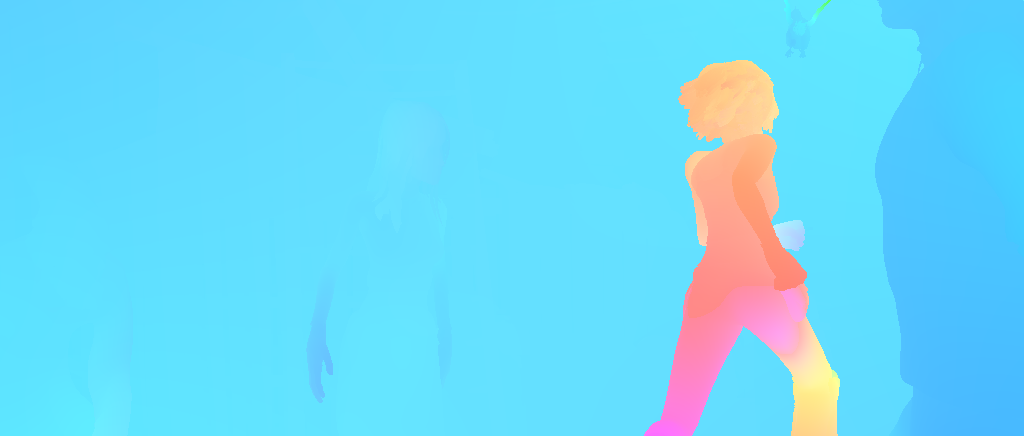}
\put(60,0){\framebox(25,20){}}
\end{overpic} \\ [-1pt]
\raisebox{20pt}{DiscreteFlow} &
\begin{overpic}[width=\imw]{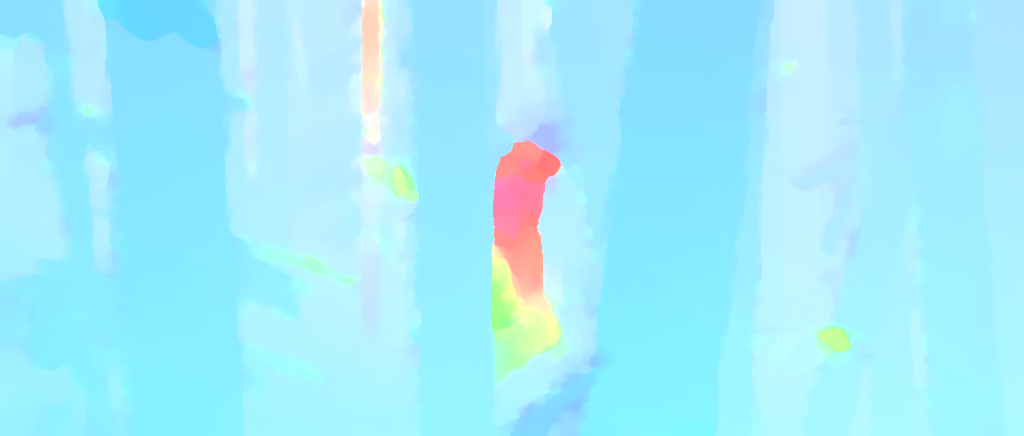}
\put(45,5){\framebox(15,35){}}
\end{overpic} &
\begin{overpic}[width=\imw]{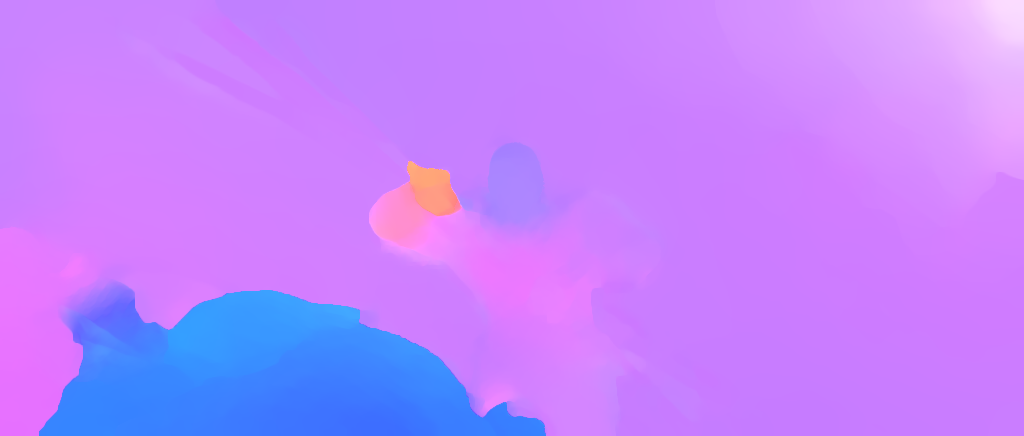}
\put(25,20){\framebox(20,20){}}
\end{overpic} &
\begin{overpic}[width=\imw]{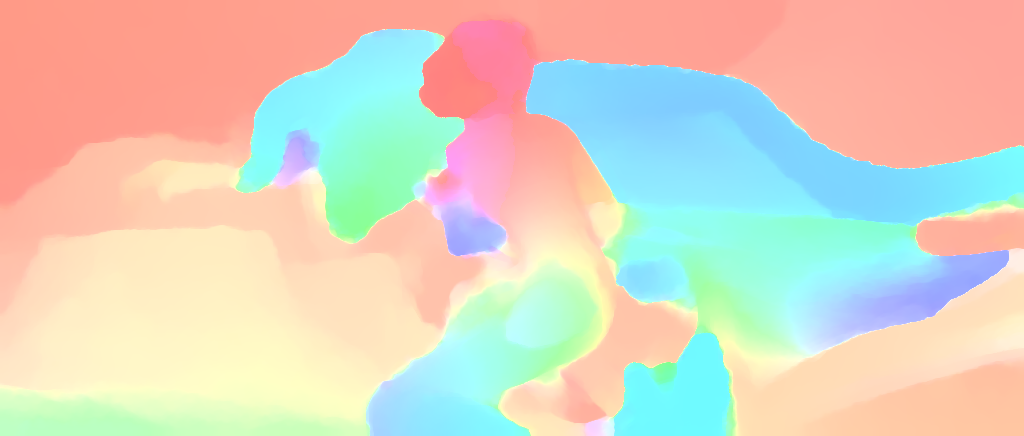}
\put(40,0){\framebox(20,20){}}
\end{overpic} &
\begin{overpic}[width=\imw]{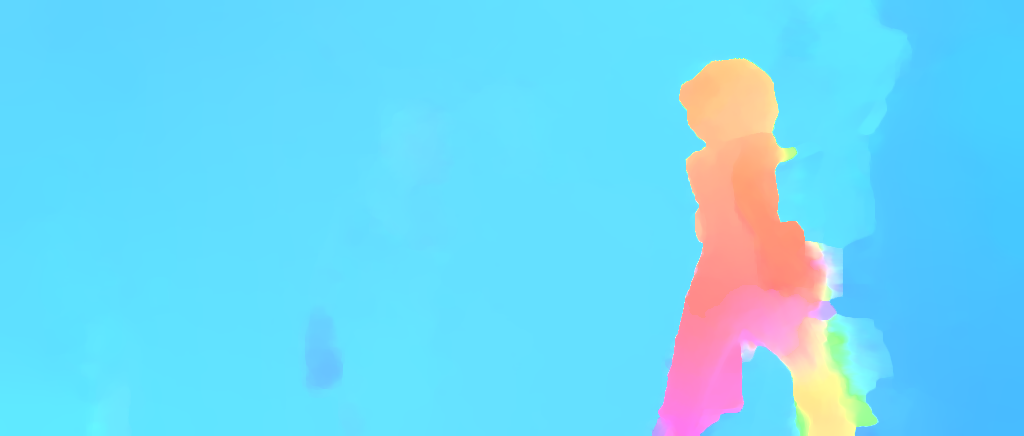}
\put(60,0){\framebox(25,20){}}
\end{overpic} \\ [-1pt]
\raisebox{20pt}{EpicFlow} &
\begin{overpic}[width=\imw]{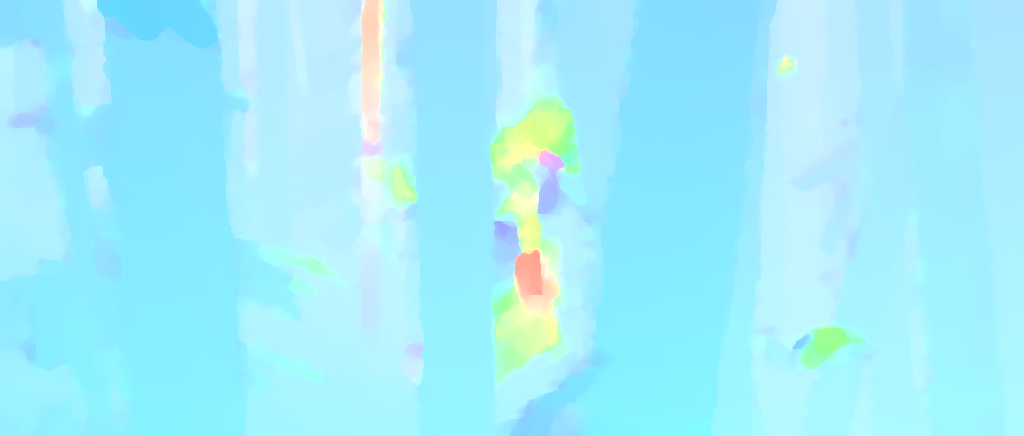}
\put(45,5){\framebox(15,35){}}
\end{overpic} &
\begin{overpic}[width=\imw]{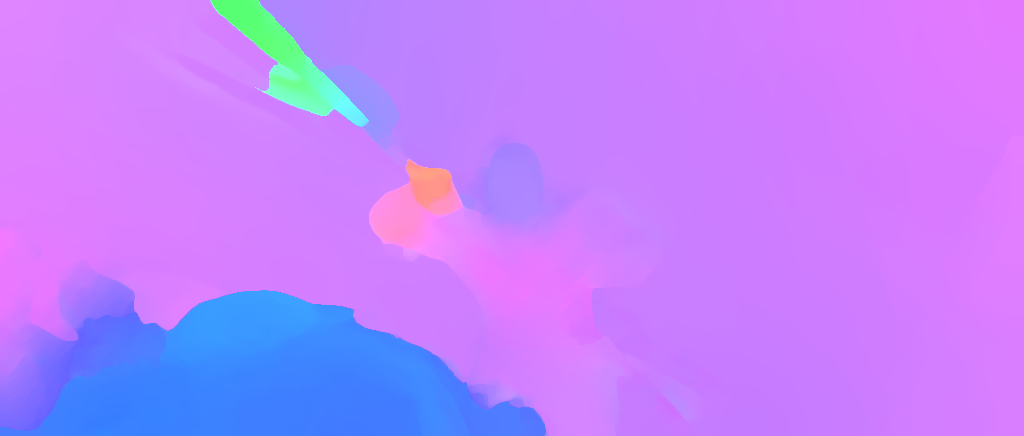}
\put(25,20){\framebox(20,20){}}
\end{overpic} &
\begin{overpic}[width=\imw]{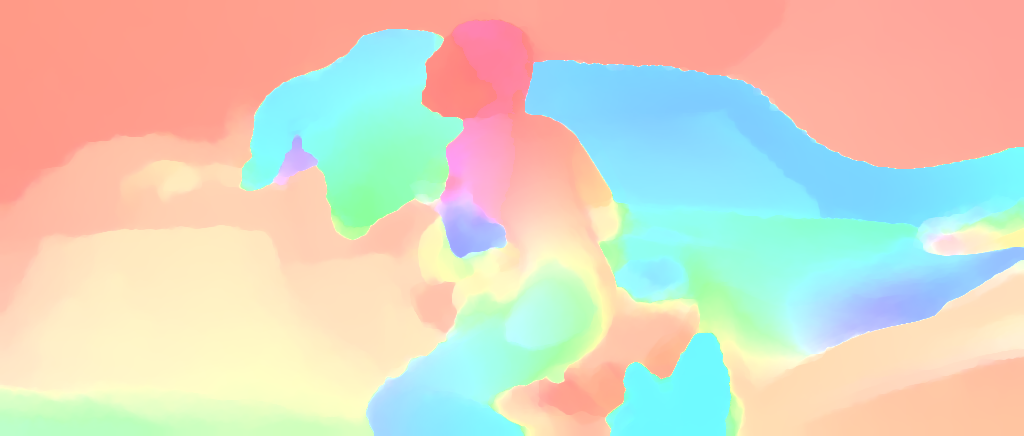}
\put(40,0){\framebox(20,20){}}
\end{overpic} &
\begin{overpic}[width=\imw]{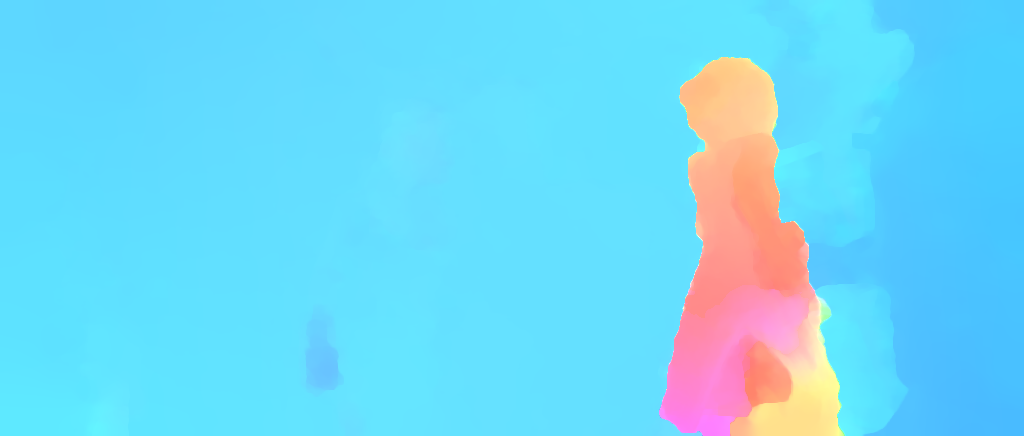}
\put(60,0){\framebox(25,20){}}
\end{overpic} \\ [-1pt]
\raisebox{20pt}{Flowfields} &
\begin{overpic}[width=\imw]{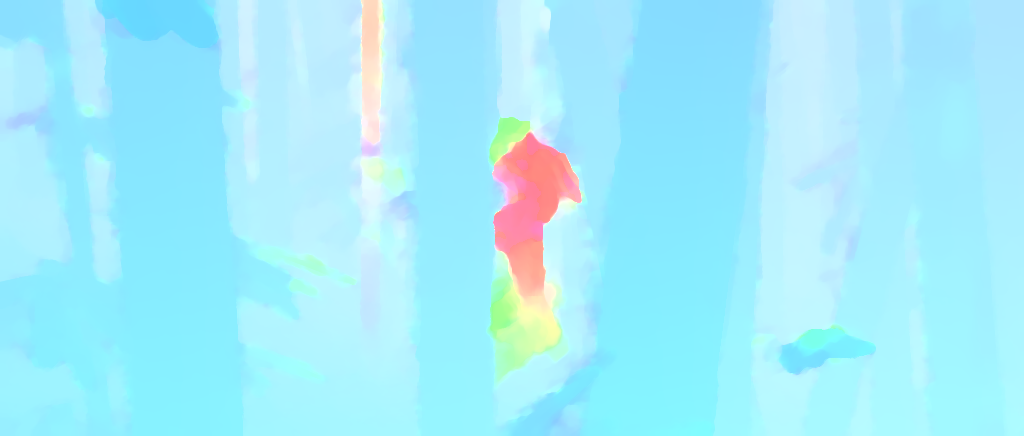}
\put(45,5){\framebox(15,35){}}
\end{overpic} &
\begin{overpic}[width=\imw]{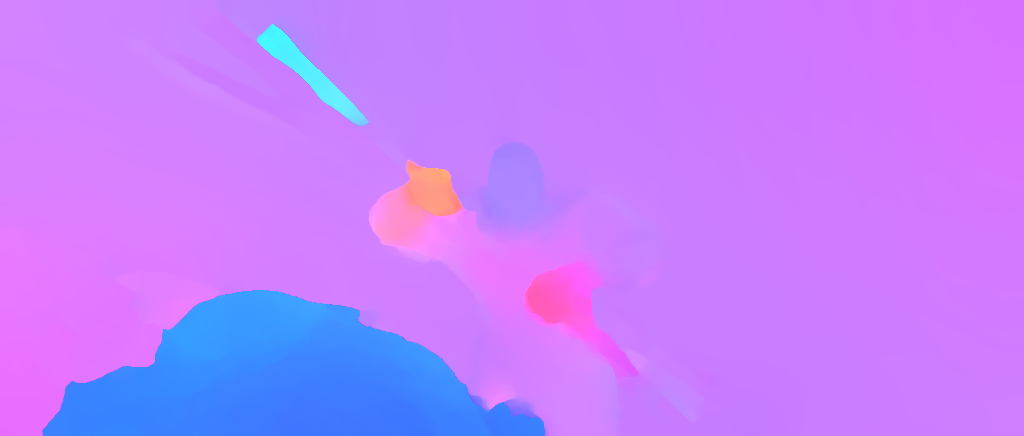}
\put(25,20){\framebox(20,20){}}
\end{overpic} &
\begin{overpic}[width=\imw]{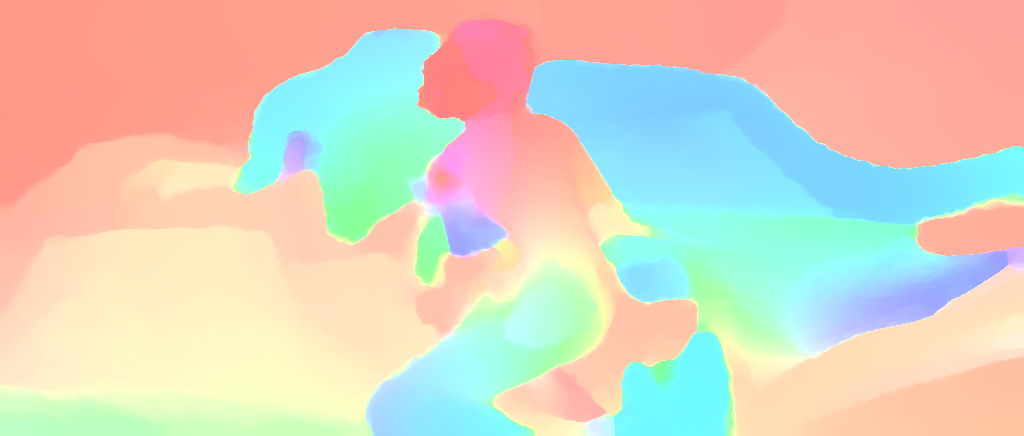}
\put(40,0){\framebox(20,20){}}
\end{overpic} &
\begin{overpic}[width=\imw]{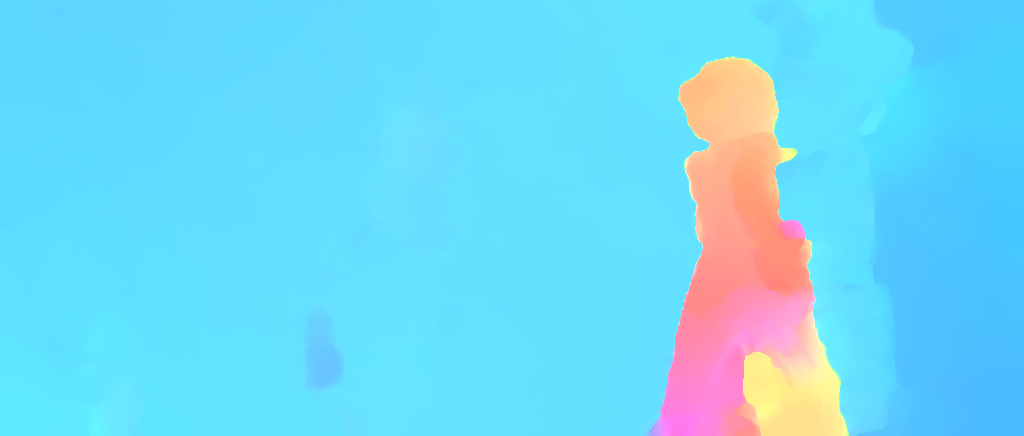}
\put(60,0){\framebox(25,20){}}
\end{overpic} \\ [-1pt]
\raisebox{20pt}{Ours} &
\begin{overpic}[width=\imw]{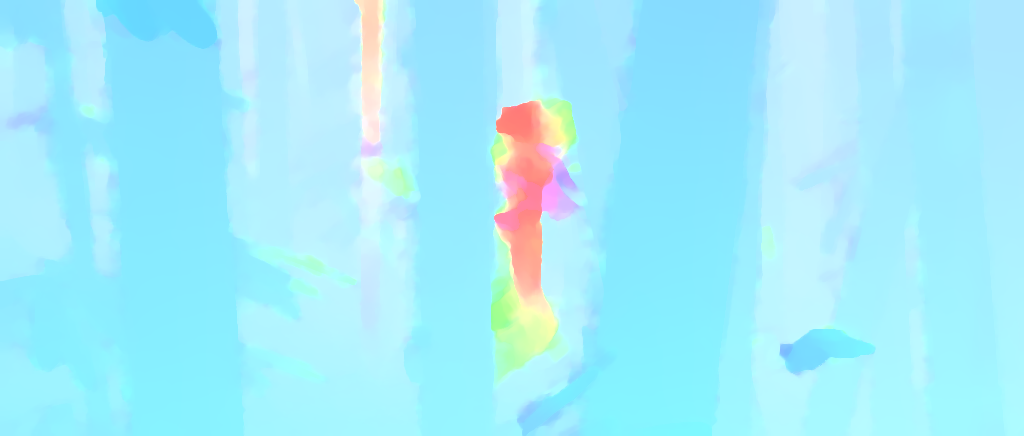}
\put(45,5){\framebox(15,35){}}
\end{overpic} &
\begin{overpic}[width=\imw]{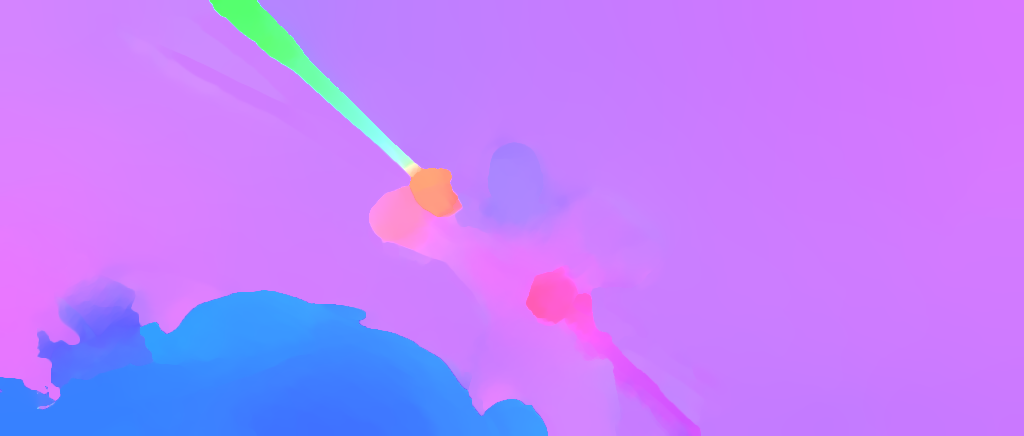}
\put(25,20){\framebox(20,20){}}
\end{overpic} &
\begin{overpic}[width=\imw]{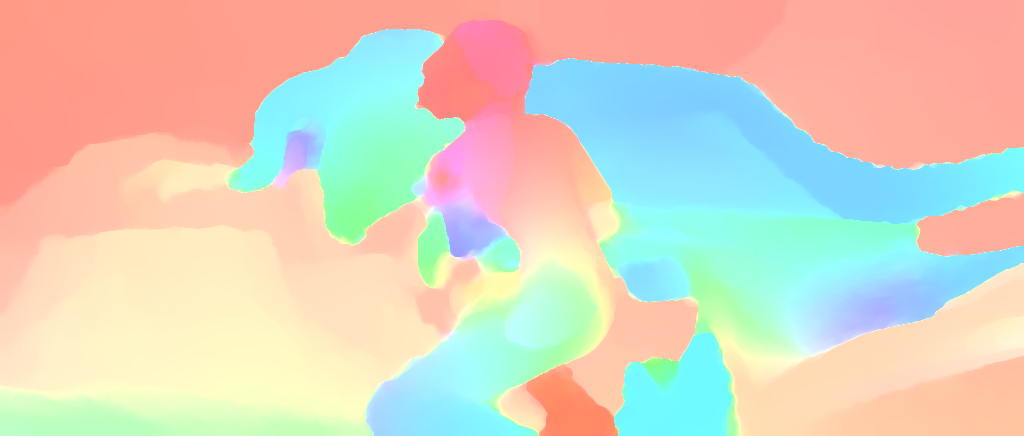}
\put(40,0){\framebox(20,20){}}
\end{overpic} &
\begin{overpic}[width=\imw]{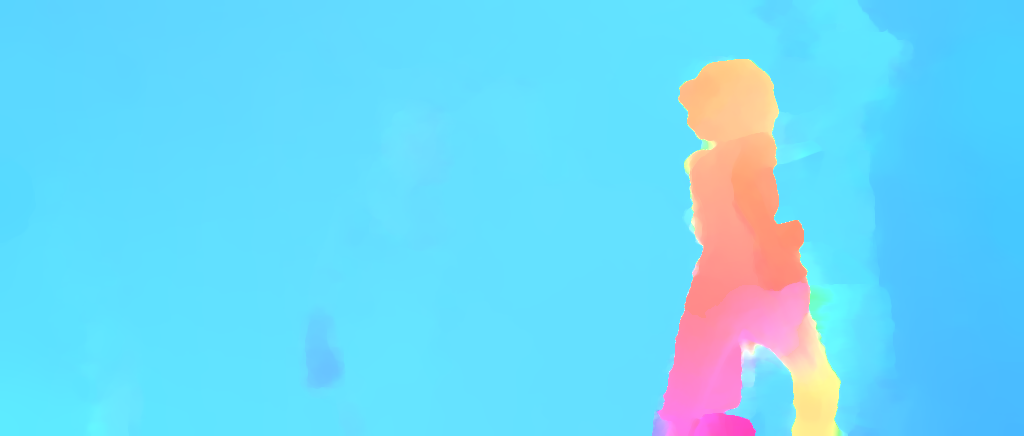}
\put(60,0){\framebox(25,20){}}
\end{overpic} \\
\end{tabular}
 \caption{Qualitative results on Sintel optical flow. Our method recovers precise motion of fine structures, like the butterfly, pole and legs as highlighted in boxes.}
\label{fig:quali_sintel}
 \end{figure*}

\subsection{Discussions} 

\paragraph{Receptive field size} The advantage of the proposed model is its content-aware receptive field size. The attention model adjusts the receptive field according to the image content through weighting each scale. Given an image pyramid with smallest scale $\times 4$, our algorithm is able to produce a maximum receptive field with $23\times 4 = 132$. This approach introduces more context into the local matching scheme. It would greatly help resolve matching ambiguity because of repetitive, smooth textures, or matching along edges. On the other hand, in regions with unique structures, our attention model learns to focus on finer scales with a relatively smaller receptive field, excluding unnecessary context to influence matching. 

\paragraph{Extensions to hand-crafted features} Our AutoScalar model can be extended to hand-crafted features, such as SIFT and DAISY. To be specific, instead of using a neural network to compute multi-scale features, we can generate multi-scale features through changing
the hyper-parameters of SIFT and DAISY. Then an attention net is trained to combine these multi-scale features in a content-aware manner towards a better performance. 

%% file: exp.tex
\section{Experiments}

\begin{figure*}
\begin{center}
\begin{subfigure}[t]{0.24\textwidth}
 \includegraphics[width=\textwidth]{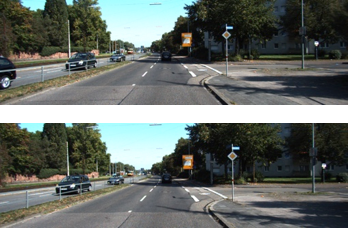}
 \caption{Input image}
 \end{subfigure}
\begin{subfigure}[t]{0.24\textwidth}
 \includegraphics[width=\textwidth]{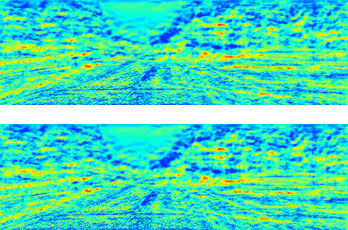}
 \caption{Dense feature slice}
 \end{subfigure}
\begin{subfigure}[t]{0.24\textwidth}
 \includegraphics[width=\textwidth]{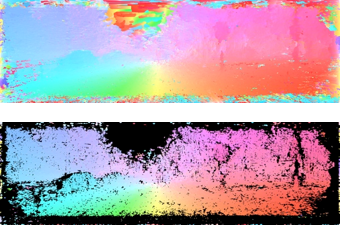}
 \caption{Init match and outlier removal}
 \end{subfigure}
\begin{subfigure}[t]{0.24\textwidth}
 \raisebox{8mm}{\includegraphics[width=\textwidth]{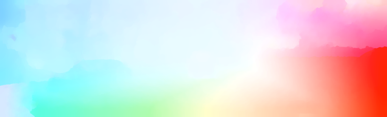}}
 \caption{Final flow estimation}
 \end{subfigure}
\end{center}
  \vspace{-0.5cm}
\caption{The optical flow estimation pipeline. From left to right: input image pairs, scale-attended feature maps, initial noisy estimation (top) and flow field after outlier removal (bottom), final result after extrapolation \cite{epicflow}. }
  \label{fig:interp}
  \vspace{-3mm}
\end{figure*}

This section presents the result of the proposed scale-attention network on both geometric matching and semantic matching tasks. For geometric matching, we perform evaluations on the challenging optical flow benchmarks, MPI-Sintel \cite{sintel} and KITTI \cite{KITTI}. The semantic matching experiment is conducted over the Caltech-UCSD Birds 2011 dataset \cite{cub}. We compare with the current state-of-the-art algorithms. Apart from the quantitative experiments, we also visualize and discuss the interpretability of the attention maps that our model generates for different tasks.

\begin{table}[t]
\begin{center}
\begin{tabu}{c|ccc} \hline
Method & EPE-matched & EPE-un & EPE-all \\ \hline
\textbf{AutoScaler} &	\textbf{2.569} &	34.656	& 6.076 \\
DeepDisFlow \cite{deepdiscreteflow} & 2.623 &	31.042 & \textbf{5.728}	\\
FlowFields\cite{flowfields} &	2.621 &	 31.799 &	5.810 \\
FullFlow \cite{fullflow} &	2.684	& \textbf{30.793} & 	5.895 \\
DiscreteFlow \cite{discreteflow} & 2.937 & 31.685 & 6.077 \\
PatchCollider \cite{gpc}	& 2.938	& 31.309	& 6.040 \\
EpicFlow \cite{epicflow} &	3.060 &	32.564 &	6.285 \\
DeepFlow2 \cite{deepflow} &	3.093 &	38.166	& 6.928 \\
FGI \cite{fgi}	& 3.101 &	35.158	 & 6.607 \\ \hline
\end{tabu}
\end{center}
\vspace{-5mm}
\caption{Quantitative experiments on Sintel Dataset.}
\label{tab:sintel-flow}
\vspace{-2mm}
\end{table}%

\subsection{Optical flow on MPI-Sintel}

We first evaluate our method on the challenging MPI-Sintel optical flow benchmark \cite{sintel}, which consists of more than 1200 pairs of training images and 1500 pairs of testing images. It is a synthetic dataset with extremely large motion from both cameras and objects with various appearance changes due to motion blur, illumination and non-rigid deformation. The benchmark error metric is end-point-error (EPE), which is the average euclidean distance between the flow fields. We refer to EPE-matched and EPE-unmatched as average end-point-error over regions that remain visible in adjacent frames and average end-point-error over regions that are visible only in one of two adjacent frames. And EPE-all is the end-point-error over all the pixels. 

\paragraph{Training data} We firstly split the 22 training images into training (1-16) and validation (17-22). For each pair of images, we randomly sampled 10K local correspondent pairs, and for each pair, we randomly selected 200 negative samples within the limit of motion range $[-210, 200]$.

\paragraph{Architecture design} In order to validate the efficacy of our proposed network, we evaluate Top-1 accuracy matching performance over baseline network architecture. We compared different architectures trained under the same multi-class siamese configuration with softmax loss. Table.~\ref{tab:top1} demonstrates the top-1 accuracy matching performance of different architectures on the validation subset of Sintel and KITTI. 
The competing algorithms include CNN-31x31, a nine-layer fully convolutional network used in \cite{deepstereo}. Similar to our approach, \cite{deepstereo} also adopts softmax loss for training and the architecture does not include pooling or stride convolution. 
``Single'' refers to our basic single-scale deep architecture with five ResNet blocks. ``Concat x2'' is a two-scale deep architecture, with the feature vector as a concatenation of features from the two scales.
{``AutoScaler $\times K$'' is the proposed AutoScaler with $K$ scales. In this experiment we compare the performance between two and four scales.} As shown in the table, our proposed architecture outperform all the competing algorithms. Especially, we show that with the attention mechanism the matching performance is better than simply concatenating two scales. Moreover, the four-scale AutoScaler outperforms the two-scale version. It is also worth noting that we extract DAISY features from multiple scales, and trained our attention network to fuse the features. We found that it outperforms concatenating multi-scale DAISY features with a large margin, which demonstrates the efficacy of the attention model. 

\paragraph{Dense Flow}\label{sec:sintel}
In the testing stage, in order to generate the dense flow, we firstly use our network to extract dense features. For each local feature from the source image, we compute the inner-product over all the local features from the target image within the motion range limit $[-240, 240]\times [-240, 240]$ and pick the highest score. This would produce a raw dense flow fields with outliers. We remove outliers through forward-backward consistency check. To be specific, for each pixel $\bp$, we check whether the condition
$\| \mathbf{u}_\mathrm{backward}(\mathbf{p} + \mathbf{u}_\mathrm{forward}(\bp)) + \mathbf{u}_\mathrm{forward}(\mathbf{p})  \| \leq t$ is satisfied, where $\mathbf{u}_\mathrm{backward}$ is the estimated backward optical flow, $\mathbf{u}_\mathrm{forward}$ is the forward optical flow field. We then discard inconsistent motion estimations with above the threshold $t$. In practice $t = 3$ is used for dense flow.
This gives us an optical flow map with partial observation. And we interpolate/extrapolate the missing pixels with Epicflow algorithm \cite{epicflow}. Fig.~\ref{fig:interp} illustrates the whole process of dense optical flow pipeline. 

\paragraph{Quantitative Results} We submit our algorithm's output to Sintel benchmark and compare it against the top-ranked prior work. We focus on the final benchmark, which is more challenging due to the presence of motion blur and various shading and reflectance changes. Table.~\ref{tab:sintel-flow} shows the quantitative results against the competing algorithms. To be specific, our method achieves the best performance on the EPE-matched measure among all the competing algorithms, and ranks 5th in EPE-all metric. This suggests the proposed network is very competitive in finding existed correspondence. However, relative large errors appear in unmatched regions, which suggests the interpolation might not be tuned to fit for our network's output. It is worth to note that, unlike some competing algorithms, such as Flowfields \cite{flowfields}, DiscreteFlow \cite{deepdiscreteflow,discreteflow} and Fullflow \cite{fullflow}, the proposed dense optical flow algorithm does not exploit a comprehensive MRF post-processing step to propagate the flow estimation to occluded regions. This would speed-up the optical flow matching process since MRFs processing is the bottleneck for many methods. In practice, our method takes 0.5s for computing features, 2 mins for initial matching and 2.5s for Epicflow interpolation on Sintel. 
\paragraph{Qualitative Results} Fig.~\ref{fig:quali_sintel} demonstrates more qualitative results for visual comparison. Thanks to the scale-attention scheme, our method has the best capability in capturing small objects with large motion, as shown in the figure. This is because our method has both large receptive field and sub-pixel resolution.
\begin{figure}
\begin{center}
\adjincludegraphics[width=0.75\linewidth, trim={{.24\width} {.375\width} {.24\width} {.41\width}},clip]{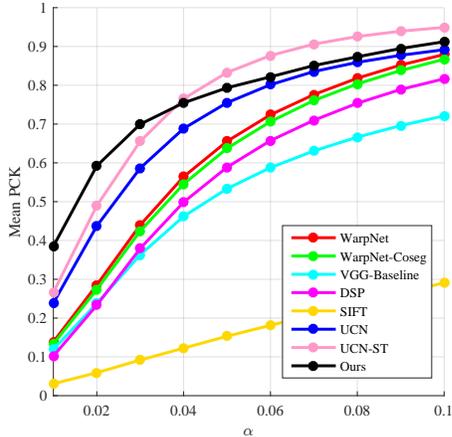}
\end{center}
  \vspace{-0.8cm}
\caption{Quantitative result on CUB dataset.}
  \label{fig:quant_cub}
  \vspace{-5mm}
\end{figure}

\subsection{Optical flow on KITTI}

We also report the benchmarking result over KITTI Optical Flow 2015 dataset \cite{KITTI}. This benchmark includes 200 image pairs for training and 200 image pairs for testing. 

\paragraph{Training data} We separate the training dataset into 160 pairs as train and 40 pairs as validation. Following the similar experiment configuration on Sintel, we sample 10k local correspondences from each image pair and for each pair 200 negative samples. 
\paragraph{Dense flow} We follow similar pipeline described in Sec.~\ref{sec:sintel} to generate dense optical flow with our network, as shown in Fig.~\ref{fig:interp}. To be specific, we firstly compute features with the proposed AutoScalar network, and conduct initial matching. Forward-backward consistency check is used to remove outliers and Epic flow is exploited to generate the final dense optical flow field. 
\paragraph{Architecture comparison}  We first report the top-1 accuracy over validation set with different architectures, as shown in Table.~\ref{tab:top1}. The proposed AutoScalar model outperforms all competing network architecture. And the attention mechanism also brings limited improvement on DAISY feature. This validates the effectiveness of the proposed architecture. {It is worth noting that on KITTI AutoScaler x2 outperforms AutoScaler x4. This contrasts what we find in Sintel. We suspect that it is because the dataset bias: KITTI does not have many large regions without any textures.}
\begin{figure*}
\centering
\begin{subfigure}[t]{0.16\textwidth}
 \includegraphics[width=\textwidth]{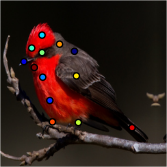} \\
\includegraphics[width=\textwidth]{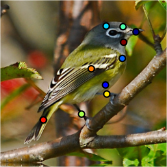} \\
  \includegraphics[width=\textwidth]{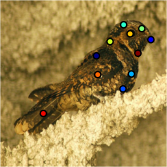} \\
 \includegraphics[width=\textwidth]{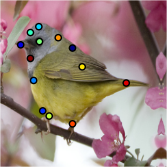}
    \vspace{-5mm}
 \caption{Query image}
 \end{subfigure}
 \begin{subfigure}[t]{0.16\linewidth}
 \includegraphics[width=\textwidth]{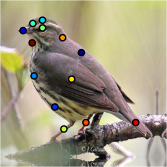} \\
\includegraphics[width=\textwidth]{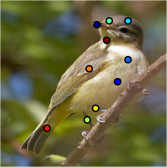} \\
  \includegraphics[width=\textwidth]{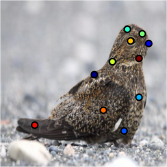} \\
 \includegraphics[width=\textwidth]{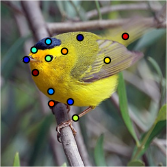}
    \vspace{-5mm}
  \caption{Ground-truth}
  \end{subfigure}
 \begin{subfigure}[t]{0.16\textwidth}
 \includegraphics[width=\textwidth]{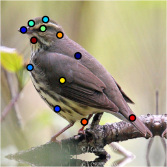} \\
\includegraphics[width=\textwidth]{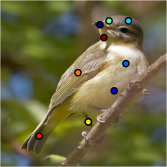} \\
  \includegraphics[width=\textwidth]{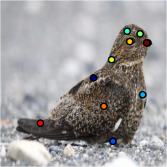} \\
 \includegraphics[width=\textwidth]{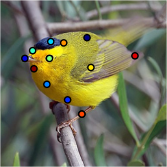}
    \vspace{-5mm}
  \caption{Ours}
 \end{subfigure}
\begin{subfigure}[t]{0.16\textwidth}
 \includegraphics[width=\textwidth]{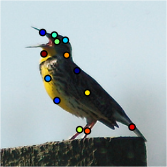} \\
 \includegraphics[width=\textwidth]{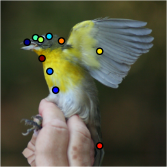} \\
  \includegraphics[width=\textwidth]{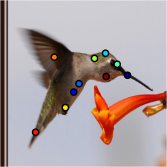} \\
 \includegraphics[width=\textwidth]{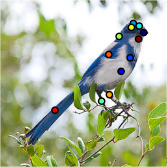}
    \vspace{-5mm}
  \caption{Query image}
 \end{subfigure}
 \begin{subfigure}[t]{0.16\linewidth}
 \includegraphics[width=\textwidth]{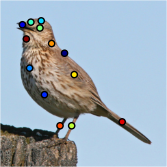} \\
 \includegraphics[width=\textwidth]{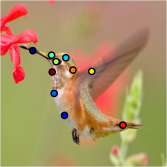} \\
  \includegraphics[width=\textwidth]{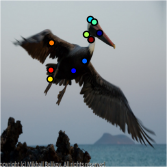} \\
 \includegraphics[width=\textwidth]{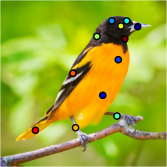}
    \vspace{-5mm}
  \caption{Ground-truth}
  \end{subfigure}
 \begin{subfigure}[t]{0.16\textwidth}
 \includegraphics[width=\textwidth]{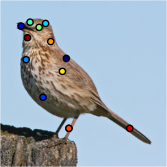} \\
 \includegraphics[width=\textwidth]{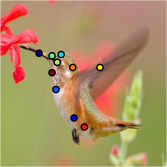} \\
  \includegraphics[width=\textwidth]{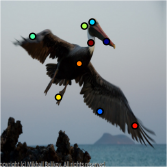} \\
 \includegraphics[width=\textwidth]{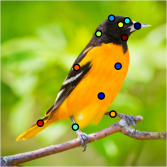}
   \vspace{-5mm}
  \caption{Ours}
 \end{subfigure}
 \vspace{-3mm}
 \caption{Qualitative results on CUB semantic matching. Our method is able to capture semantic meaningful matching across species and poses, with sub-pixel level accuracy. A typical failure case is the left-right feet ambiguity (see the bottom row). }
 \end{figure*}
\paragraph{Quantitative results} We submit our result to KITTI optical flow benchmark. 
The results are shown in Table.~\ref{tab:kitti-flow}. Note that KITTI is a dataset captured in a special autonomous driving scenario, where the motion is mainly due to the ego-motion of the camera plus rigid motion of the cars in the scene. Thus dense flow approaches that exploit the semantics of the scene objects as well as the epipolar constraint would achieve significant improvement \cite{sdf, sof}. Apart from those methods, our approach achieves comparable results against other competing algorithms utilizes generic matching techniques. 
From the table we can see our method is comparable with most competing algorithms. The proposed method is not favorable among all the deep learning based algorithms. One potential reason is the proposed method does not exploit the extrapolation, which brings large error in non-visible regions because of self-occlusion and truncation. We plan to incorporate structured variational prediction into the proposed model to solve this problem in future.

\begin{table}[t]
\begin{center}
\begin{tabu}{c|ccc} \hline
Method & Fl-bg & Fl-fg & Fl-all \\ \hline
SDF \cite{sdf} & \textbf{8.61 \%} &	\textbf{26.69} \% & \textbf{11.62 \%}	\\
SOF \cite{sof} & 14.63 \% & 27.73 \% & 16.81 \% \\ \hline
CNN-HPM\cite{cnn-hpm} & {18.33} \% & {24.96 \%} & {19.44}\% \\
DiscreteFlow \cite{discreteflow} & 21.53 \% & 26.68 \% & 22.38 \% \\
\textbf{AutoScaler} &	21.85 \% & 31.62 \% & 25.64 \% \\
FullFlow \cite{fullflow} &	23.09 \% & 30.11 \% & 24.26 \% \\
EpicFlow \cite{epicflow} &	25.81 \% & 33.56 \% & 27.10 \% \\
DeepFlow2 \cite{deepflow} &	27.96 \% & 35.28 \% & 29.18 \% \\
PatchCollider \cite{gpc}	& 30.60 \% & 33.09 \% & 31.01 \% \\
SGM+C+NL \cite{classicnl}	& 40.81 \% & 35.42 \% & 39.91 \% \\ \hline
\end{tabu}
\end{center}
\vspace{-5mm}
\caption{Quantitative experiments on KITTI Flow 2015 Dataset. The metrics for KITTI benchmark 'Fl-bg', 'Fl-fg' and 'Fl-all' represent the outlier percentage on background pixels, foreground pixels and all pixels respectively.}
\label{tab:kitti-flow}
\vspace{-2mm}
\end{table}%


\subsection{Semantic Matching}

Unlike geometric matching tasks, such as optical flow and stereo, the semantic matching aims at finding correspondence that represents coherent semantic meanings, regardless whether these keypoints are similar in appearance, \etc. 
We perform the semantic matching experiments on the CUB-2011-2011 dataset, which contains 11788 images of 200 bird categories, with 15 parts annotated. 

\paragraph{Training data} We follow the experiment configuration of \cite{warpnet}, which utilizes the training set to extract training pairs and 5000 pairs images from the validation subset as testing pairs. We crop each image with the bounding box of the bird and conduct matching on the cropped image pairs. For each training iteration, we randomly pick two pairs of images and use all the corresponding keypoints between them for training. The negative samples are randomly selected over the whole target images. 

\paragraph{Metric} We evaluate the accuracy of matches with the percentage of correct keypoints (PCK@$\alpha$). A match is considered as correct if it lies with $\alpha L$ pixels of the ground-truth correspondence, where $L = \frac{1}{2} (\sqrt{w_{\mathrm{src}}^2 + h_{\mathrm{src}}^2} + \sqrt{w_{\mathrm{tgt}}^2 + h_{\mathrm{tgt}}^2}) $ is the mean diagonal size of the image pairs. Note that not all the 15 keypoints are visible in both images, we follow the configuration of \cite{warpnet} and discard these invisible keypoints when computing the metric. 

\paragraph{Quantitative result} We compared against the more recent state-of-the-arts algorithms on CUB matching dataset, namely WarpNet \cite{warpnet}, Universal correspondence network \cite{ucn}, and DSP \cite{dsp}, along with two widely used features including VGGnet \cite{vgg} and SIFT \cite{sift}. Fig.~\ref{fig:quant_cub} depicts the PCK metric along different threshold $\alpha$. From this figure we can see that our method outperforms all the competing algorithms when $\alpha$ is small, which suggests the highest subpixel accuracy. When the threshold $\alpha$ becomes large, our method ranks second among all the competing algorithm, following UCN \cite{ucn}. This suggests that AutoScaler better captures finer accurate details while in the meantime performs competitively in reasoning the semantic meaning of the local part of the birds. Fig.~\ref{fig:quant_cub} show the examples of the qualitative matching results. As shown in this figure, our method performs well in most cases across various poses, species and scales. The most failure cases are due to the ambiguity in matching left and right feet. 


%% file: conclusion.tex
\section{Conclusions}

We propose the AutoScaler, a scale-attention network that optimally combines dense feature maps from different scales. This scheme allows our neural network to have an adaptive receptive field size. The extensive experiments show that our method is not only extremely effective but is also able to generate visual interpretable scale attentions.  

%% file: main.bbl
\begin{thebibliography}{10}\itemsep=-1pt

\bibitem{rome}
S.~Agarwal, N.~Snavely, I.~Simon, S.~M. Seitz, and R.~Szeliski.
\newblock Building rome in a day.
\newblock In {\em ICCV}, 2009.

\bibitem{ba-attention-2015}
J.~Ba, V.~Mnih, and K.~Kavukcuoglu.
\newblock Multiple object recognition with visual attention.
\newblock In {\em ICLR}, 2015.

\bibitem{sdf}
M.~Bai, W.~Luo, K.~Kundu, and R.~Urtasun.
\newblock Exploiting semantic information and deep matching for optical flow.
\newblock In {\em ECCV}, 2016.

\bibitem{flowfields}
C.~Bailer, B.~Taetz, and D.~Stricker.
\newblock Flow fields: Dense correspondence fields for highly accurate large
  displacement optical flow estimation.
\newblock In {\em ICCV}, 2015.

\bibitem{cnn-hpm}
C.~Bailer, K.~Varanasi, and D.~Stricker.
\newblock Cnn based patch matching for optical flow with thresholded hinge
  loss.
\newblock {\em arXiv}, 2016.

\bibitem{PatchMatch}
C.~Barnes, E.~Shechtman, D.~B. Goldman, and A.~Finkelstein.
\newblock The generalized patchmatch correspondence algorithm.
\newblock In {\em ECCV}. 2010.

\bibitem{surf}
H.~Bay, T.~Tuytelaars, and L.~Van~Gool.
\newblock Surf: Speeded up robust features.
\newblock In {\em ECCV}, 2006.

\bibitem{sintel}
D.~J. Butler, J.~Wulff, G.~B. Stanley, and M.~J. Black.
\newblock A naturalistic open source movie for optical flow evaluation.
\newblock In {\em ECCV}, 2012.

\bibitem{chen}
L.-C. Chen, Y.~Yang, J.~Wang, W.~Xu, and A.~L. Yuille.
\newblock Attention to scale: Scale-aware semantic image segmentation.
\newblock {\em CVPR}, 2016.

\bibitem{fullflow}
Q.~Chen and V.~Koltun.
\newblock Full flow: Optical flow estimation by global optimization over
  regular grids.
\newblock In {\em CVPR}, 2016.

\bibitem{baidu-deep}
Z.~Chen, X.~Sun, L.~Wang, Y.~Yu, and C.~Huang.
\newblock A deep visual correspondence embedding model for stereo matching
  costs.
\newblock In {\em ICCV}, 2015.

\bibitem{universalnet}
C.~B. Choy, J.~Gawk, S.~Savarese, and M.~Chandraker.
\newblock Universal correspondence network.
\newblock {\em NIPS}, 2016.

\bibitem{ucn}
C.~B. Choy, J.~Gwak, S.~Savarese, and M.~Chandraker.
\newblock Universal correspondence network.
\newblock In {\em NIPS}, 2016.

\bibitem{fusion4d}
M.~Dou, S.~Khamis, Y.~Degtyarev, P.~Davidson, S.~R. Fanello, A.~Kowdle, S.~O.
  Escolano, C.~Rhemann, D.~Kim, J.~Taylor, et~al.
\newblock Fusion4d: real-time performance capture of challenging scenes.
\newblock {\em SIGGRAPH}, 2016.

\bibitem{BroxMalik}
B.~Drayer and T.~Brox.
\newblock Combinatorial regularization of descriptor matching for optical flow
  estimation.
\newblock In {\em BMVC}, 2015.

\bibitem{KITTI}
A.~Geiger, P.~Lenz, and R.~Urtasun.
\newblock Are we ready for autonomous driving? the kitti vision benchmark
  suite.
\newblock In {\em CVPR}, 2012.

\bibitem{rcnn}
R.~Girshick, J.~Donahue, T.~Darrell, and J.~Malik.
\newblock Rich feature hierarchies for accurate object detection and semantic
  segmentation.
\newblock In {\em CVPR}, 2014.

\bibitem{draw15}
K.~Gregor, I.~Danihelka, A.~Graves, D.~Rezende, and D.~Wierstra.
\newblock Draw: A recurrent neural network for image generation.
\newblock In {\em ICML}. 2015.

\bibitem{videoseg}
M.~Grundmann, V.~Kwatra, M.~Han, and I.~Essa.
\newblock Efficient hierarchical graph-based video segmentation.
\newblock In {\em Computer Vision and Pattern Recognition (CVPR), 2010 IEEE
  Conference on}, pages 2141--2148. IEEE, 2010.

\bibitem{deepdiscreteflow}
F.~G{\"u}ney and A.~Geiger.
\newblock Deep discrete flow.
\newblock In {\em ACCV}, 2016.

\bibitem{matchnet}
X.~Han, T.~Leung, Y.~Jia, R.~Sukthankar, and A.~C. Berg.
\newblock Matchnet: Unifying feature and metric learning for patch-based
  matching.
\newblock In {\em Proceedings of the IEEE Conference on Computer Vision and
  Pattern Recognition}, pages 3279--3286, 2015.

\bibitem{dsp}
T.~Harada, Y.~Ushiku, Y.~Yamashita, and Y.~Kuniyoshi.
\newblock Discriminative spatial pyramid.
\newblock In {\em CVPR}, 2011.

\bibitem{sifting}
T.~Hassner, S.~Filosof, V.~Mayzels, and L.~Zelnik-Manor.
\newblock Sifting through scales.
\newblock {\em PAMI}, 2016.

\bibitem{sls}
T.~Hassner, V.~Mayzels, and L.~Zelnik-Manor.
\newblock On sifts and their scales.
\newblock In {\em Computer Vision and Pattern Recognition (CVPR)}. IEEE, 2012.

\bibitem{resnet}
K.~He, X.~Zhang, S.~Ren, and J.~Sun.
\newblock Deep residual learning for image recognition.
\newblock {\em CVPR}, 2016.

\bibitem{warpnet}
A.~Kanazawa, D.~W. Jacobs, and M.~Chandraker.
\newblock Warpnet: Weakly supervised matching for single-view reconstruction.
\newblock {\em CVPR}, 2016.

\bibitem{libviso2}
B.~Kitt, A.~Geiger, and H.~Lategahn.
\newblock Visual odometry based on stereo image sequences with ransac-based
  outlier rejection scheme.
\newblock In {\em IV}, 2010.

\bibitem{alexnet}
A.~Krizhevsky, I.~Sutskever, and G.~E. Hinton.
\newblock Imagenet classification with deep convolutional neural networks.
\newblock In {\em Advances in neural information processing systems}, pages
  1097--1105, 2012.

\bibitem{fgi}
Y.~Li, D.~Min, M.~N. Do, and J.~Lu.
\newblock Fast guided global interpolation for depth and motion.
\newblock In {\em ECCV}, 2016.

\bibitem{scale-selection}
T.~Lindeberg.
\newblock Feature detection with automatic scale selection.
\newblock {\em IJCV}, 1998.

\bibitem{fcn}
J.~Long, E.~Shelhamer, and T.~Darrell.
\newblock Fully convolutional networks for semantic segmentation.
\newblock In {\em CVPR}, 2015.

\bibitem{ning}
J.~L. Long, N.~Zhang, and T.~Darrell.
\newblock Do convnets learn correspondence?
\newblock In {\em Advances in Neural Information Processing Systems}, pages
  1601--1609, 2014.

\bibitem{sift}
D.~G. Lowe.
\newblock Object recognition from local scale-invariant features.
\newblock In {\em ICCV}, 1999.

\bibitem{deepstereo}
W.~Luo, A.~G. Schwing, and R.~Urtasun.
\newblock Efficient deep learning for stereo matching.
\newblock In {\em CVPR}, 2016.

\bibitem{mser}
J.~Matas, O.~Chum, M.~Urban, and T.~Pajdla.
\newblock Robust wide-baseline stereo from maximally stable extremal regions.
\newblock {\em IVC}, 2004.

\bibitem{discreteflow}
M.~Menze, C.~Heipke, and A.~Geiger.
\newblock Discrete optimization for optical flow.
\newblock In {\em GCPR}, 2015.

\bibitem{affine-region}
K.~Mikolajczyk, T.~Tuytelaars, C.~Schmid, A.~Zisserman, J.~Matas,
  F.~Schaffalitzky, T.~Kadir, and L.~V. Gool.
\newblock A comparison of affine region detectors.
\newblock {\em Int. J. Comput. Vision}, 65(1-2):43--72, Nov. 2005.

\bibitem{attention_nips14}
V.~Mnih, N.~Heess, A.~Graves, and k.~kavukcuoglu.
\newblock Recurrent models of visual attention.
\newblock In {\em NIPS}. 2014.

\bibitem{orbslam}
R.~Mur-Artal, J.~Montiel, and J.~D. Tard{\'o}s.
\newblock Orb-slam: a versatile and accurate monocular slam system.
\newblock {\em IEEE Trans on Robotics}, 2015.

\bibitem{dtam}
R.~A. Newcombe, S.~J. Lovegrove, and A.~J. Davison.
\newblock Dtam: Dense tracking and mapping in real-time.
\newblock In {\em ICCV}, 2011.

\bibitem{scalesift}
W.~Qiu, X.~Wang, X.~Bai, A.~Yuille, and Z.~Tu.
\newblock Scale-space sift flow.
\newblock In {\em WACV}, 2014.

\bibitem{fasterrcnn}
S.~Ren, K.~He, R.~Girshick, and J.~Sun.
\newblock Faster r-cnn: Towards real-time object detection with region proposal
  networks.
\newblock In {\em NIPS}, 2015.

\bibitem{epicflow}
J.~Revaud, P.~Weinzaepfel, Z.~Harchaoui, and C.~Schmid.
\newblock Epicflow: Edge-preserving interpolation of correspondences for
  optical flow.
\newblock In {\em CVPR}, 2015.

\bibitem{foeflow}
S.~Roth and M.~J. Black.
\newblock On the spatial statistics of optical flow.
\newblock In {\em IJCV}, 2007.

\bibitem{sof}
L.~Sevilla-Lara, D.~Sun, V.~Jampani, and M.~J. Black.
\newblock Optical flow with semantic segmentation and localized layers.
\newblock In {\em CVPR}, 2016.

\bibitem{convex}
K.~Simonyan, A.~Vedaldi, and A.~Zisserman.
\newblock Learning local feature descriptors using convex optimisation.
\newblock {\em IEEE Transactions on Pattern Analysis and Machine Intelligence},
  36(8):1573--1585, 2014.

\bibitem{vgg}
K.~Simonyan and A.~Zisserman.
\newblock Very deep convolutional networks for large-scale image recognition.
\newblock {\em arXiv preprint arXiv:1409.1556}, 2014.

\bibitem{SGM}
D.~Sun, S.~Roth, and M.~J. Black.
\newblock Secrets of optical flow estimation and their principles.
\newblock In {\em CVPR}, 2010.

\bibitem{classicnl}
D.~Sun, S.~Roth, and M.~J. Black.
\newblock Secrets of optical flow estimation and their principles.
\newblock In {\em CVPR}, 2010.

\bibitem{googlenet}
C.~Szegedy, W.~Liu, Y.~Jia, P.~Sermanet, S.~Reed, D.~Anguelov, D.~Erhan,
  V.~Vanhoucke, and A.~Rabinovich.
\newblock Going deeper with convolutions.
\newblock In {\em CVPR}, 2015.

\bibitem{daisy}
E.~Tola, V.~Lepetit, and P.~Fua.
\newblock Daisy: An efficient dense descriptor applied to wide-baseline stereo.
\newblock In {\em PAMI}, 2010.

\bibitem{boosting}
T.~Trzcinski, M.~Christoudias, V.~Lepetit, and P.~Fua.
\newblock Learning image descriptors with the boosting-trick.
\newblock In {\em Advances in neural information processing systems}, pages
  269--277, 2012.

\bibitem{cub}
C.~Wah, S.~Branson, P.~Welinder, P.~Perona, and S.~Belongie.
\newblock The caltech-ucsd birds-200-2011 dataset.
\newblock 2011.

\bibitem{gpc}
S.~Wang, S.~Ryan~Fanello, C.~Rhemann, S.~Izadi, and P.~Kohli.
\newblock The global patch collider.
\newblock In {\em CVPR}, 2016.

\bibitem{deepflow}
P.~Weinzaepfel, J.~Revaud, Z.~Harchaoui, and C.~Schmid.
\newblock Deepflow: Large displacement optical flow with deep matching.
\newblock In {\em ICCV}, 2013.

\bibitem{attention_icml15}
K.~Xu, J.~Ba, R.~Kiros, K.~Cho, A.~Courville, R.~Salakhutidinov, R.~Zemel, and
  Y.~Bengio.
\newblock Show, attend and tell: Neural image caption generation with visual
  attention.
\newblock In {\em ICML}. 2015.

\bibitem{dilation}
F.~Yu and V.~Koltun.
\newblock Multi-scale context aggregation by dilated convolutions.
\newblock {\em ICLR}, 2016.

\bibitem{siamese}
S.~Zagoruyko and N.~Komodakis.
\newblock Learning to compare image patches via convolutional neural networks.
\newblock In {\em CVPR}, 2015.

\bibitem{mccnn}
J.~{\v{Z}}bontar and Y.~LeCun.
\newblock Computing the stereo matching cost with a convolutional neural
  network.
\newblock In {\em CVPR}, 2015.

\end{thebibliography}
